\documentclass[conference,a4paper]{IEEEtran}
\usepackage{graphicx}
\usepackage{amsmath}
\usepackage{amssymb}
\usepackage{booktabs}
\usepackage[hyphens]{url}
\usepackage[colorlinks=true,allcolors=blue]{hyperref}
\usepackage{algpseudocode}
\usepackage{algorithm}
\usepackage{tabularx}
\usepackage{multirow}
\usepackage{float}
\usepackage{microtype} % Optimizes character spacing to prevent text overflow
\usepackage{orcidlink}

\title{Multi-Agent Off-Policy Deep Reinforcement Learning for Smart Campus Coverage}

\author{
    \IEEEauthorblockN{Omar Rady\orcidlink{0009-0003-3350-2330}, Mohamed Ayman\orcidlink{0009-0004-0134-5817}, Ali Arafa\orcidlink{0009-0008-2342-9324}, Mohamed Shalma\orcidlink{0000-0002-7185-1393}}
    \IEEEauthorblockA{Faculty of Information Engineering and Technology\\
    The German University in Cairo\\
    Cairo, Egypt\\
    }
}

\begin{document}
\maketitle

\begin{abstract}
Deep reinforcement learning (DRL) has recently gained a great attention due to its real-time adaptation and effectiveness in complex optimization problems. This paper investigates the optimal deployment of millimeter-wave (mmWave) base stations (BSs) in a realistic, non-convex campus topology. The optimization problem is NP-hard, due to the non-convex, non-smooth nature of the max-min fairness objective. To overcome these constraints, we formulate the BS placement as a Markov Decision Process (MDP) and systematically benchmark four DRL schemes: a discrete single-agent Deep Q-Network (DQN), a spatially partitioned Multi-Agent DQN, a continuous single-agent Deep Deterministic Policy Gradient (DDPG), and a geographically partitioned multi-agent DDPG framework. Numerical evaluations reveal that the multi-agent DDPG approach substantially outperforms single-agent in dense scenarios. Additionally full coverage is achieved, and a fairness Jain's index of 0.94 is obtained. Finally, the multi-agent demonstrates highly efficient computational convergence of dense scenarios with $400$ users.
\end{abstract}


\section{Introduction}
Next-generation (B5G/6G) wireless networks increasingly depend on the millimeter-wave (mmWave) spectrum to accommodate explosive demands for massive connectivity and multi-gigabit throughput. However, the high-frequency nature of mmWave transmissions makes them highly vulnerable to physical blockages, severe penetration attenuation, and atmospheric absorption. Consequently, defining the exact spatial coordinates for base station (BS) infrastructure is no longer a standard planning exercise; it constitutes an NP-hard optimization challenge that dictates network reliability, spectral efficiency, and equitable user fairness.

In practical urban and campus environments, deployment terrains rarely adhere to symmetric or convex geometries, which limits the signal coverage \cite{10525785}. Conventional mathematical solvers, such as mixed-integer non-linear programming (MINLP) or exhaustive grid-search algorithms, struggle to resolve these non-convex spatial constraints. They either incur prohibitive computational overhead or fail to capture the highly directional propagation dynamics of mmWave signals. To overcome these limitations, Deep Reinforcement Learning (DRL) gained increased attention as a robust, model-free alternative, enabling autonomous agents to iteratively learn and refine optimal solutions \cite{11232082,11313425}.

While recent literature has applied DRL and heuristic techniques to mmWave propagation and dynamic load balancing \cite{10.1177/1550147720926374, 333, 10.1145/3573942.3574002,10942890}, many existing models rely on simplified, unobstructed environments or predefined candidate coordinates \cite{222}. For example, while DQN can jointly optimize mmWave coverage and localization \cite{al2024multiobjective}, such frameworks often rely on discrete grid representations that restrict action spaces to predefined movements \cite{al2024multiobjective}. Conversely, although continuous actor-critic DRL frameworks enable flexible 3D placement of multiple UAV base stations to maximize user Quality of Experience (QoE) \cite{hoang2023adaptive}, they assume unrestricted aerial mobility \cite{hoang2023adaptive}. The vast majority of the above-mentioned works are limited to single-BS scenarios and single agent modeling. As the environment becomes more complex, the curse of dimensionality dominates leading to a notable degradation in the DRL agents performance. There is a lack of unified optimization frameworks that simultaneously address mmWave characteristics, multi-agent modeling, and the strict topological constraints of non-convex topologies. The primary contributions of this work are as follows:
\begin{itemize}
\item We address the joint optimization of multiple base stations placement for a multi-user scenario to enhance the minimum rate for maximum fairness using mmWaves in a practical urban environment.
\item The considered BS placement vicinity is a multi-space non-convex topology that leads to a non-convex and consequently, an NP-hard optimization problem.
\item We propose the use of both continuous DRL approaches like DDPG and discrete ones as DQN. Furthermore, the two approaches are evaluated using singe and multi-agent architectures. By introducing a boundary projection mechanism and a rigorous max-min fairness objective, the multi-agent DDPG framework surpasses other models and achieves superior optimality, guaranteeing $100\%$ coverage, absolute minimum SNRs above $19$ dB, and a Jain's index of 0.94.
\end{itemize}

\section{System and Channel Model}
\subsection{Model Description}
\begin{figure}[!t]
    \centering
    \includegraphics[width=0.8\linewidth]{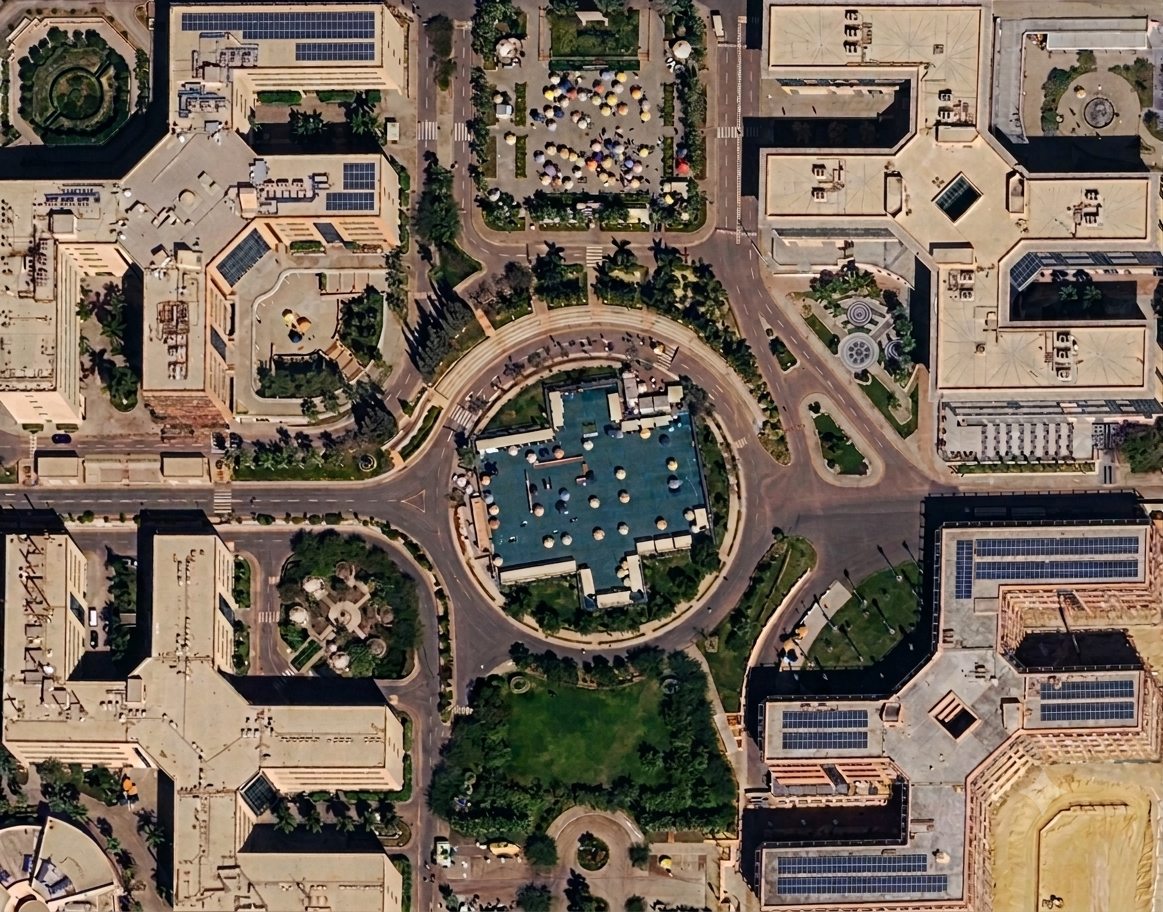}
    \caption{Aerial view of the full campus environment at the German University in Cairo (GUC) showing all main buildings.}
    \label{fig:campus}
\end{figure}

We analyze a downlink mmWave communication architecture where BSs are strategically deployed to serve $K$ stationary user equipments (UEs) distributed throughout an indoor/courtyard environment. The geometrical layout strictly models the C-Buildings complex at the Germa university in Cairo (GUC), as illustrated in Fig. \ref{fig:campus}, featuring four adjacent rectangular structures that enclose a central open courtyard. The optimization goal is to extract the ideal two-dimensional (2D) horizontal coordinates for placing one BS on each of the four building roofs, with the UEs distributed at ground level.

\subsection{Channel Model}
The Euclidean 3D distance separating a candidate base station and a user $k$ is formulated as
\begin{equation}
d_k = \sqrt{(x_{\text{BS}} - x_k)^2 + (y_{\text{BS}} - y_k)^2 + (h_{\text{BS}} - h_k)^2}
\end{equation}
where $(x_{\text{BS}}, y_{\text{BS}})$ and $(x_k, y_k)$ represent the horizontal plane coordinates of the BS and user $k$, while $h_{\text{BS}}$ and $h_k$ denote their respective heights. The mmWave channel state incorporating large-scale path loss and small-scale fading is modeled as
\begin{equation}
h_k = \sqrt{L(d_k)} \, g_k
\end{equation}
with $g_k \sim \mathcal{CN}(0,1)$ capturing Rayleigh fading. The large-scale path loss follows the standard power law:
\begin{equation}
L(d_k) = C_0 d_k^{-\alpha}
\end{equation}
where $C_0$ specifies the path loss at a 1-meter reference, and $\alpha$ is the attenuation exponent. The received signal vector at user $k$ is given by
\begin{equation}
y_k = \sqrt{P_{\text{tx}}} \, h_k x + n_k
\end{equation}
where $P_{\text{tx}}$ is the transmission power, $x$ is the unit-power data symbol, and $n_k \sim \mathcal{CN}(0, N_0)$ is the additive white Gaussian noise. In a noise-limited mmWave regime, the signal-to-noise ratio (SNR) is expressed as
\begin{equation}
\gamma_k = \frac{P_{\text{tx}} |h_k|^2}{N_0}
\end{equation}
Yielding an achievable data rate for user $k$ of:
\begin{equation}
R_k = \log_2 \left(1 + \gamma_k \right)
\end{equation}
\begin{figure}[!t]
    \centering
    \includegraphics[width=1.1\linewidth]{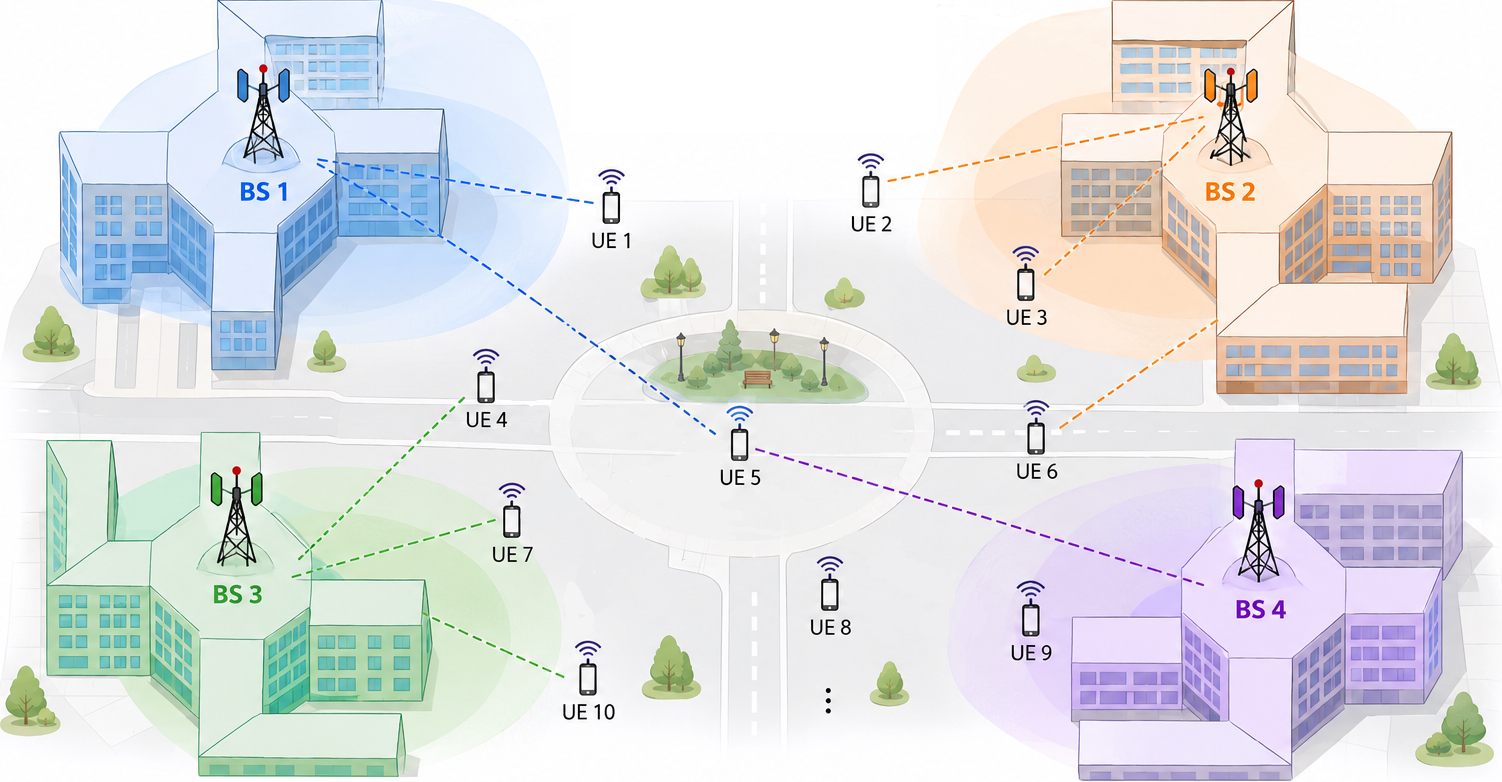}
    \caption{System model showing the 4 BSs and multiple users}
    \label{fig:placeholder}
\end{figure}
\section{Proposed DRL-Based BS Placement}
\subsection{MDP Formulation}
The proposed DRL architecture is defined by an MDP tuple encompassing state observations, continuous/discrete action spaces, and objective-driven reward signals shared across the learning agents. The DRL environments and policy networks are implemented utilizing the Stable-Baselines3 framework to ensure robust and reproducible training.
\begin{itemize}
\item \textbf{Observations}: The environmental state captures the 2D spatial distribution of all users, defined as 
\begin{equation}
    \texttt{obsv}=[(x_1,y_1),(x_2,y_2),\cdots,(x_K,y_K)]
\end{equation}
\item \textbf{Action Space}: The agents output the optimal 2D coordinates for the four deployed BSs. Therefore, the joint action vector is formalized as
\begin{equation}
    a_t=[x_{\text{BS,,1}}, y_{\text{BS,1}}, \dots, x_{\text{BS,4}}, y_{\text{BS,4}}]
\end{equation}
\item \textbf{Reward Function}: The reward mechanism directly reflects the max-min fairness target, defined strictly by the minimum user throughput rate to ensure equitable spatial coverage
\begin{equation}
    \texttt{reward} = \min_k \log_2(1+\gamma_k)
\end{equation}
\end{itemize}
\subsection{Multi-Agent DQN}

The proposed Multi-Agent Deep Q-Network (MADQN) decomposes the BS placement problem into four cooperative subproblems by partitioning the campus into four regions. A DQN agent is assigned to each region and selects the optimal BS location from a discrete set of candidate positions. Each agent maintains its own Q-network, target network, and replay buffer, enabling decentralized learning while reducing the search space. After all agents complete their selections, the minimum SNR among all campus users is evaluated and used as a shared reward, encouraging fair wireless coverage across the campus. The overall optimization procedure is summarized in Algorithm~\ref{alg:madqn}.

\begin{algorithm}[!t]
\caption{Multi-Agent DQN for BS Placement}
\label{alg:madqn}
\begin{algorithmic}[1]
\State \textbf{Input:} User locations $\{(x_k,y_k)\}_{k=1}^{K}$ and candidate BS locations for each region
\For{each agent $i=1,\ldots,4$}
    \State Initialize Q-network $Q_{\theta_i}$ and target network $Q_{\bar{\theta}_i}$
    \State Initialize replay buffer $\mathcal{D}_i$
\EndFor
\For{each episode}
    \State Reset the environment
    \For{each agent $i$}
        \State Observe local state $s_i$
        \State Select action $a_i$ using an $\epsilon$-greedy policy
        \If{$a_i$ is invalid}
            \State Apply invalid action penalty
        \EndIf
        \State Assign BS location corresponding to $a_i$
    \EndFor
    \State Compute the global reward (minimum user SNR)
    \For{each agent $i$}
        \State Store transition $(s_i,a_i,r_i,s_i')$ in $\mathcal{D}_i$
        \State Sample a minibatch from $\mathcal{D}_i$
        \State Update $Q_{\theta_i}$
        \State Update target network $Q_{\bar{\theta}_i}$
    \EndFor
\EndFor
\State \textbf{Output:} Optimal BS locations for the four campus regions
\end{algorithmic}
\end{algorithm}

\subsection{Continuous Multi-Partitioning via DDPG}
To overcome the inherent resolution limits of grid discretization and avoid local spatial optima, the multi-partition DDPG approach explicitly subdivides the non-convex topography of each of the four building roofs into $5$ distinct geometric partitions (North, South, East, West, and Center arms). Consequently, a dedicated continuous actor-critic agent is independently trained on each of these 20 total sub-sectors (Algorithm 2). Unlike the discrete DQN framework, if the DDPG network predicts a coordinate outside the bounds of its designated partition, the point is automatically projected to the nearest valid geometric boundary on the specific roof polygon. The framework then determines the absolute best deployment for the building by selecting the winning partition that yields the highest maximum reward.

\begin{algorithm}[!t]
\caption{Multi-Agent DDPG for BS Placement}
\begin{algorithmic}[1]
\State \textbf{Input:} Users $\{(x_k,y_k)\}_{k=1}^K$, buildings $b \in \{1..4\}$, partitions $p \in \{1..5\}$
\For{each building $b$}
    \For{each partition $p$}
        \State Initialize actor $\pi_{\theta_{b,p}}$ and critic $Q_{\phi_{b,p}}$
        \State Set target networks $\pi_{\bar{\theta}_{b,p}}, Q_{\bar{\phi}_{b,p}}$
        \State Initialize partitioned replay buffer $\mathcal{D}_{b,p}$
        
        \For{each episode}
            \State Initialize environment
            \For{each time step $t$}
                \State Observe state $s_t$
                \State Generate action: $a_t^{(b,p)} = \pi_{\theta_{b,p}}(s_t) + \mathcal{N}_t$
                \State Enforce partition boundary projection: $a_t^{(b,p)} \leftarrow \text{Proj}_{\mathcal{A}_{b,p}}(a_t^{(b,p)})$
                \State Apply BS placement $(x_{\text{BS}}^b,y_{\text{BS}}^b) \leftarrow a_t^{(b,p)}$
                \State Compute step reward $r_t^{(b,p)}$
                \State Observe next state $s_{t+1}$
                \State Store tuple $(s_t,a_t^{(b,p)},r_t^{(b,p)},s_{t+1})$ in $\mathcal{D}_{b,p}$
                
                \State Sample minibatch from $\mathcal{D}_{b,p}$
                \State Update critic network using Bellman equation
                \State Ascend policy gradient to update actor
                \State Perform soft target updates: $\bar{\theta}_{b,p} \leftarrow \tau \theta_{b,p} + (1-\tau)\bar{\theta}_{b,p}$
            \EndFor
        \EndFor
    \EndFor
    \State Select winning partition per building: $(x_{\text{BS}}^{b*},y_{\text{BS}}^{b*}) = \arg\max_{p,t} r_t^{(b,p)}$
\EndFor

\State \textbf{Output:} Optimal continuous coordinates for all 4 buildings
\end{algorithmic}
\end{algorithm}

\section{Numerical Results}
This section benchmarks the proposed optimization configurations via simulation using the environmental parameters detailed in Table \ref{tab:sim_params}. The primary performance criteria include achievable max-min rates (bps/Hz), spectral efficiency, signal-to-noise ratio (SNR) distributions, and overall sector coverage percentages. 

\begin{table}[H]
\centering
\caption{Simulation Configuration Parameters}
\label{tab:sim_params}
\begin{tabular}{ll}
\toprule
\textbf{Parameter Description} & \textbf{Value} \\
\midrule
Total Number of Users ($K$) & $400$ ($100$ per zone) \\
Transmit Power ($P_{\text{tx}}$) & $1.0$\,W \\
Noise Floor Power ($N_0$) & $1 \times 10^{-10}$\,W \\
Path Loss Exponent ($\alpha$) & $4.0$ \\
BS Antenna Height & $12.0$\,m \\
User Equipment Height & $1.7$\,m \\
DDPG Learning Rate & $3 \times 10^{-4}$ \\
DQN Learning Rate & $3 \times 10^{-4}$ \\
Replay Buffer Capacity & $50{,}000$ \\
Temporal Discount Factor ($\gamma$) & $0.99$ \\
\bottomrule
\end{tabular}
\end{table}

\begin{table*}[t]
\centering
\caption{ Coverage of Campus and Fairness Evaluation via Jain's Index}
\label{tab:fairness_metrics}
\begin{tabular}{lcccc}
\toprule
\textbf{Optimization Technique} & \textbf{Total Users ($n$)} & \textbf{Mean Rate (bps/Hz)} & \textbf{Min SNR (dB)} & \textbf{Jain's Index ($J$)} \\
\midrule
Single-Agent DQN & $400$ & 3.17& 0.01& 0.71\\
Multi-Agent DQN & $400$ & 9.05& 14.01& 0.82\\
Single-Agent DDPG & $400$ & $9.64$ & $19.19$ & $0.94$\\
Multi-Agent DDPG & $400$ & $9.65$ & $19.18$ & $0.9486$ \\
\bottomrule
\end{tabular}
\end{table*}

\begin{figure}[htbp]
    \centering
    \includegraphics[width=\linewidth]{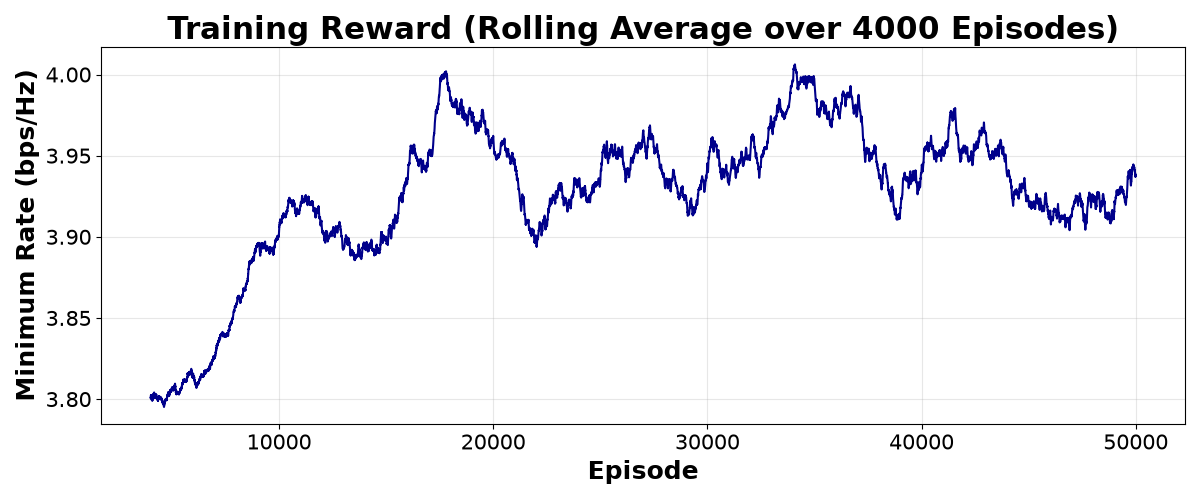}
    \caption{Training reward convergence profile for the Single-Agent DQN.}
    \label{fig:dqn_single}
\end{figure}

As depicted in Fig. \ref{fig:dqn_single}, the baseline Single-Agent DQN struggles with the scale of the centralized discrete grid. The moving average reward plateaus at approximately $3.95$ bps/Hz. This constrained performance highlights the geometric limitations of quantizing a non-convex space when attempting to optimize highly sensitive mmWave spatial coordinates.

\begin{figure}[htbp]
    \centering
    \includegraphics[width=\linewidth]{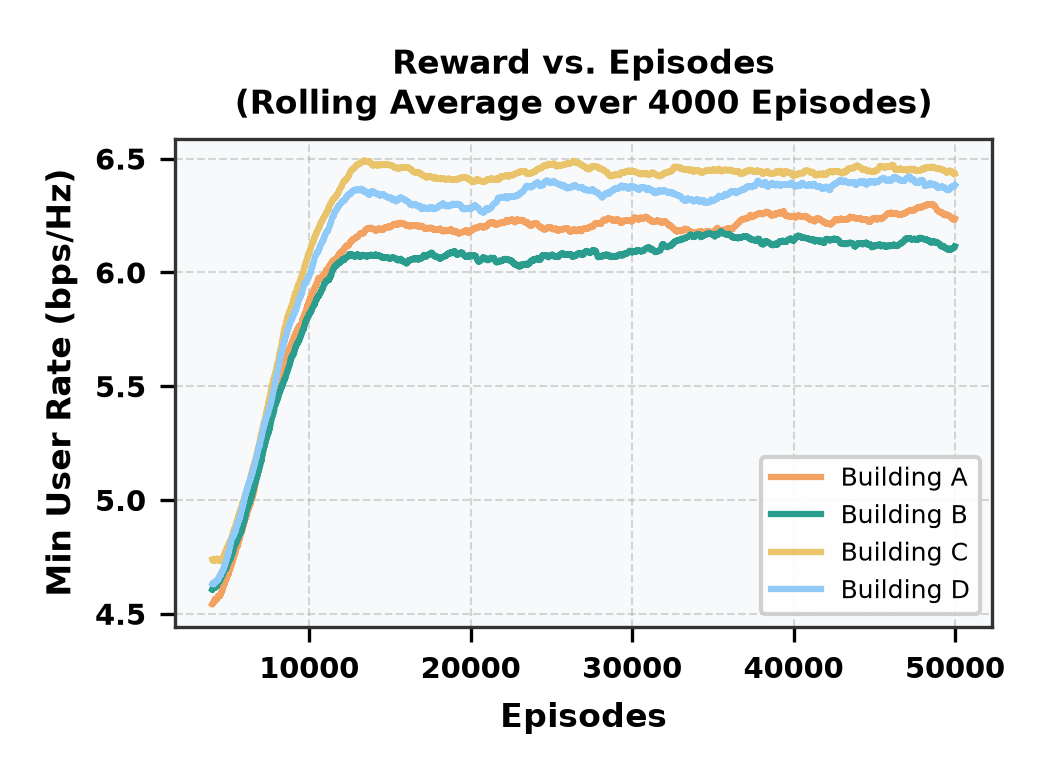}
    \caption{Training reward convergence profile for the Multi-Agent DQN.}
    \label{fig:dqn_multi}
\end{figure}

To mitigate centralized grid congestion, Fig. \ref{fig:dqn_multi} tracks the Multi-Agent DQN formulation. By assigning distinct discrete sub-grids to each of the four rooftops, the distributed agents successfully partition the computational complexity, yielding a noticeably higher and more stable max-min reward floor compared to the single-agent variant. 

\begin{figure}[htbp]
    \centering
    \includegraphics[width=\linewidth]{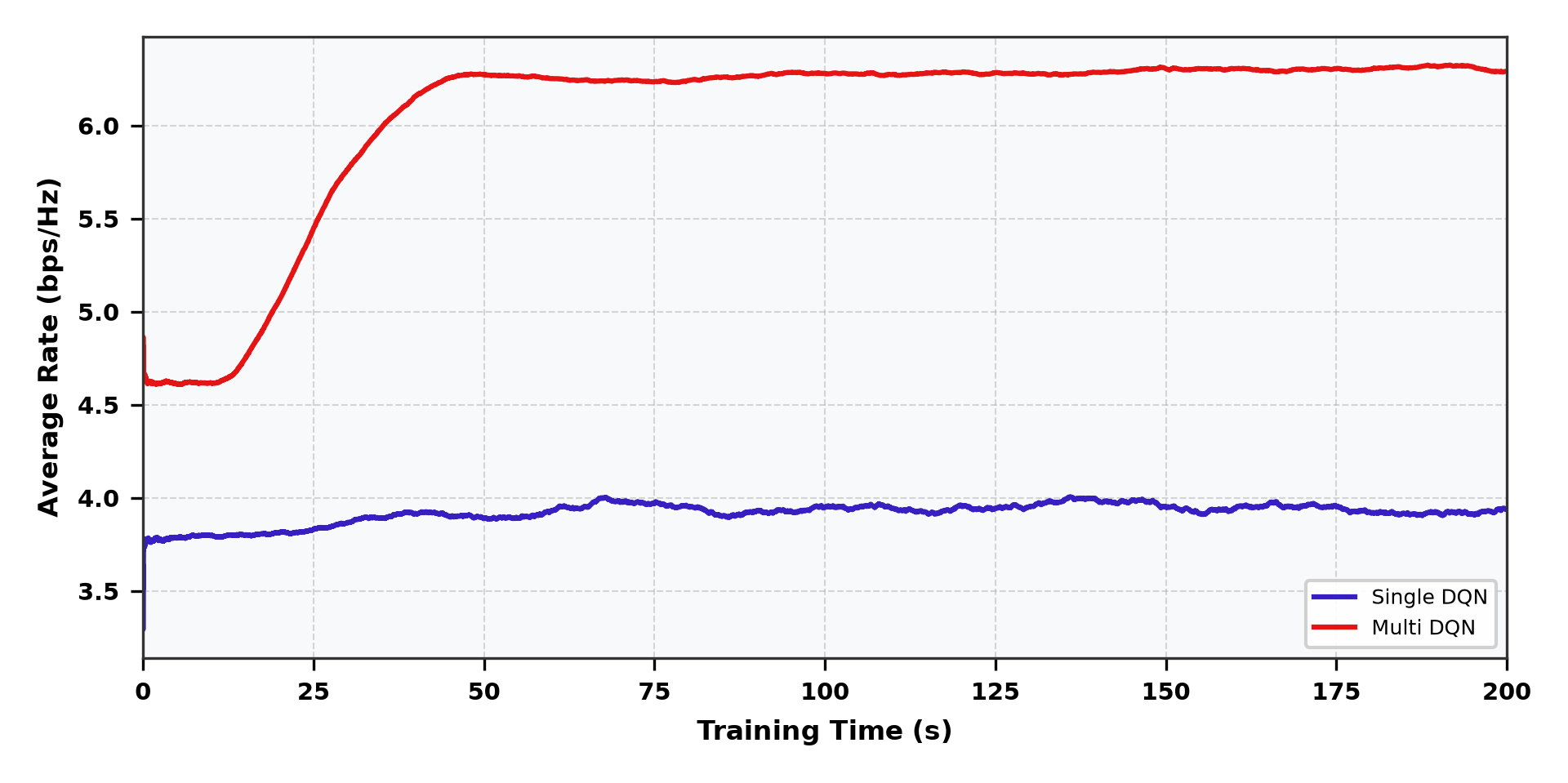}
    \caption{Time-based convergence comparison of average rates between Single-Agent and Multi-Agent DQN frameworks.}
    \label{fig:dqn_compare}
\end{figure}

Furthermore, to highlight the computational efficiency of the distributed approach, Fig. \ref{fig:dqn_compare} contrasts the average rate convergence against actual training time in seconds. The Multi-Agent DQN not only achieves a higher average rate but also demonstrates that partitioning the state space does not incur prohibitive time delays, rapidly establishing a superior performance floor compared to the single-agent baseline.

\begin{figure}[htbp]
    \centering
    \includegraphics[width=\linewidth]{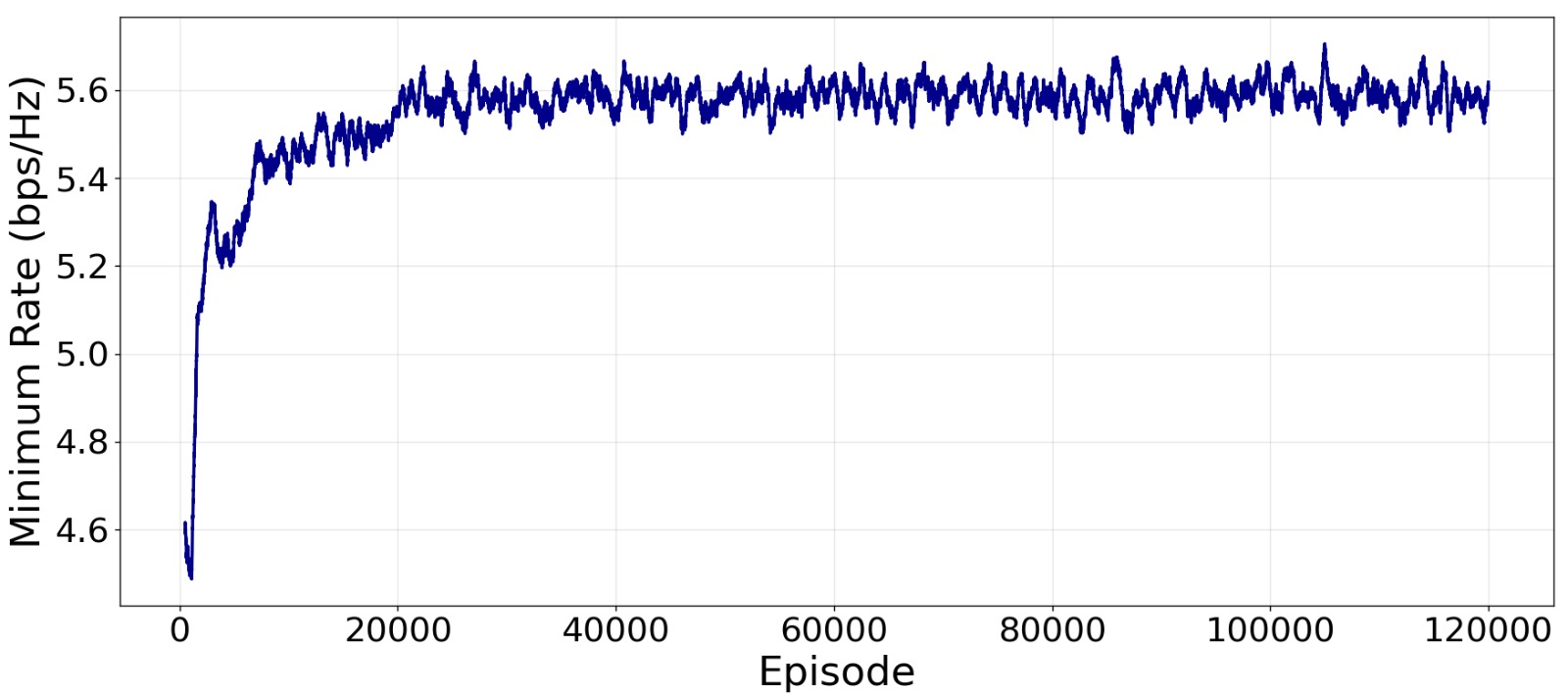}
    \caption{Episodic training reward progression for the continuous Single-Agent DDPG.}
    \label{fig:ddpg_single}
\end{figure}

Evaluating continuous action spaces, Fig. \ref{fig:ddpg_single} demonstrates the Single-Agent DDPG operating across the unbounded, un-partitioned campus geometry. By eliminating discrete grid limitations, the continuous agent achieves a substantially higher minimum rate convergence of approximately $5.6$ bps/Hz. Nevertheless, managing four concurrent coordinates via a single centralized policy still introduces variance, as the agent struggles to balance overlapping coverage zones.

\begin{figure}[htbp]
    \centering
    \includegraphics[width=\linewidth]{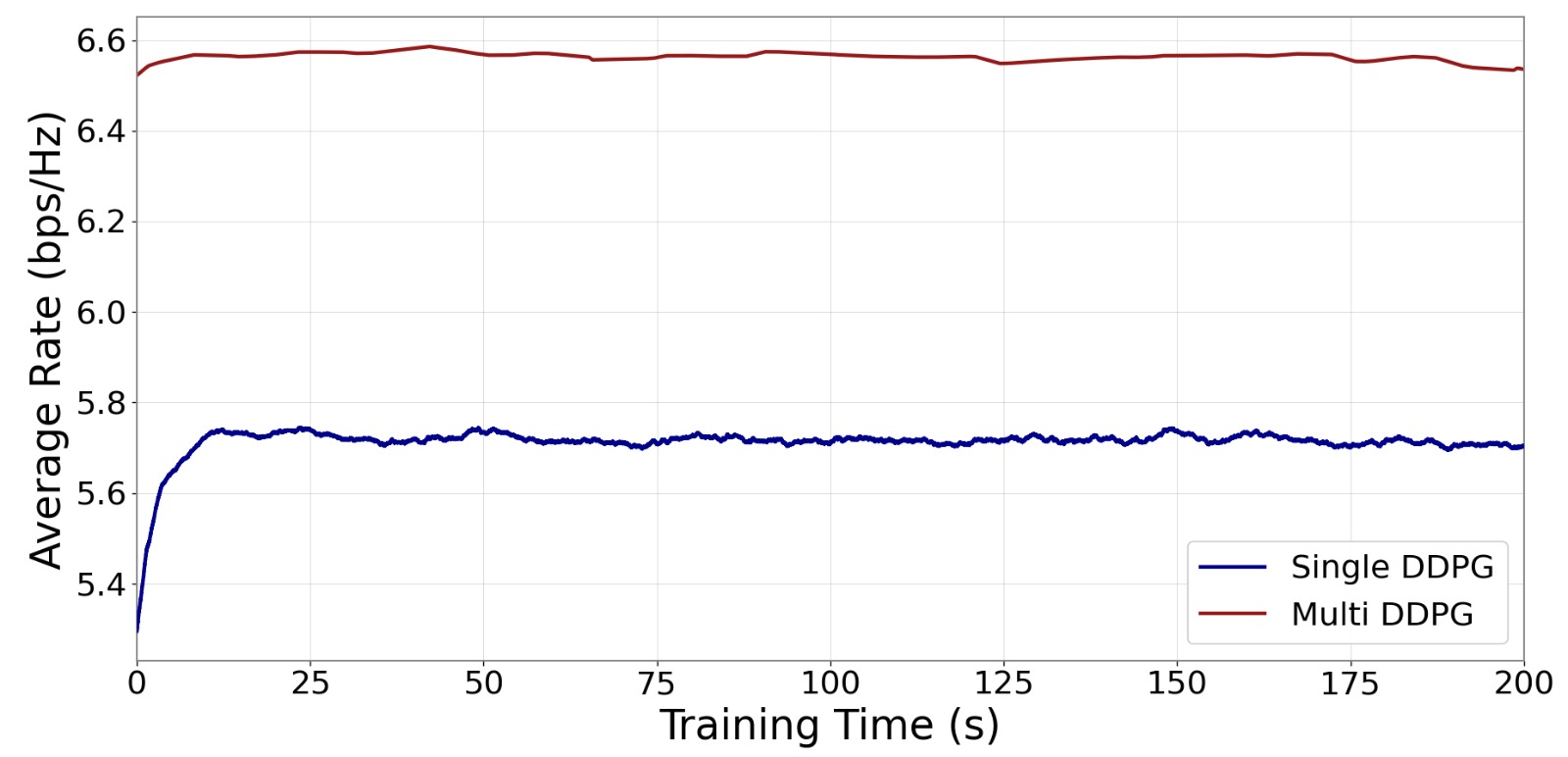}
    \caption{Time-based convergence comparison of average achievable rates for Single-Agent and Multi-Agent DDPG.}
    \label{fig:ddpg_compare}
\end{figure}

Building on the episodic performance, Fig. \ref{fig:ddpg_compare} illustrates the average rate convergence with respect to continuous training time. The multi-agent partitioning scheme maintains a highly stable average rate. By dedicating independent actors to bounded partitions, the system circumvents the extensive search times required by a single centralized agent, proving highly efficient in real-time execution.

\begin{figure}[htbp]
    \centering
    \includegraphics[width=\linewidth]{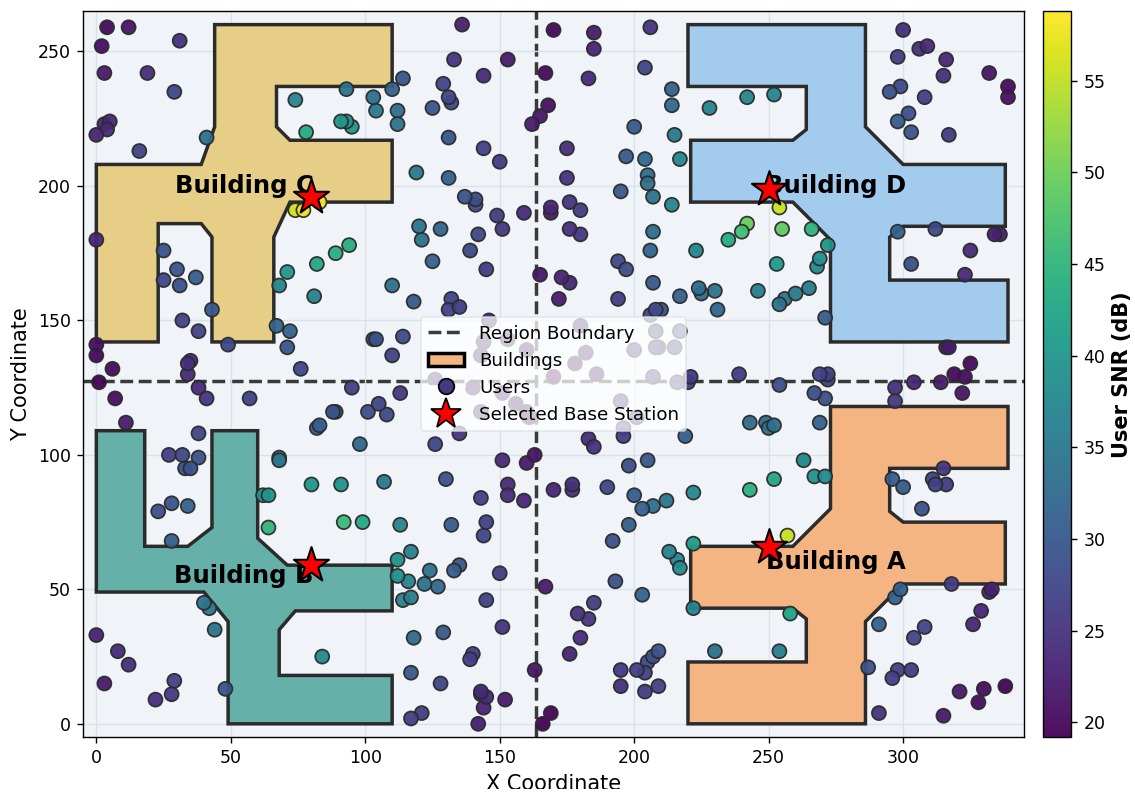}
    \caption{Optimal spatial mapping and resulting SNR distribution generated by the Multi-Agent DDPG framework.}
    \label{fig:ddpg_multi}
\end{figure}

To achieve global spatial optimality, the Multi-Agent DDPG framework deploys dedicated continuous agents to the strictly bounded rooftop partitions. Fig. \ref{fig:ddpg_multi} confirms the superiority of this localized exhaustive approach. By isolating the continuous search bounds into 5 arms per building, the framework avoids spatial minima and rapidly converges on an optimal deployment topology yielding a network-wide minimum reward of $6.393$ bps/Hz. The system achieves a flawless $100\%$ quadrant coverage ($> 0.0$ dB). Furthermore, the distributed agents deliver exceptional spectral efficiencies: $9.54$, $9.52$, $9.74$, and $9.80$ bps/Hz across Buildings A, B, C, and D, respectively, while sustaining absolute minimum SNRs above $19$ dB.

\section{Conclusion}
This study addressed the complex challenge of optimizing mmWave base station deployments across the non-convex rooftop infrastructure of the GUC campus. By formulating the environment as an MDP, we evaluated four distinct DRL methodologies targeting sum-rate and max-min fairness. Analytical results confirm that the geographically partitioned Multi-Agent DDPG approach significantly outperforms discrete grid-based and centralized continuous baselines. By dividing the complex rooftops into 5 distinct partitions and dedicating independent actor-critic agents to exhaustively search each continuous bound, the proposed framework bypasses dimensional scalability limits and local spatial minima. Ultimately, this approach achieves a superior minimum data rate of $6.393$ bps/Hz, robust SNRs exceeding $19$ dB, and perfectly equitable quadrant coverage.

\bibliographystyle{IEEEtran}
\bibliography{Ref}

\end{document}